\documentclass[11pt,a4paper]{article}
\usepackage[hyperref]{emnlp2020}
\usepackage{times}
\usepackage{latexsym}

\usepackage[inline]{enumitem}
\usepackage{graphicx}
\usepackage{amssymb}
\usepackage{hyperref}
\usepackage{arydshln}

\usepackage{multicol}
\usepackage{hyperref}

\usepackage{xcolor}

\usepackage{appendix}
\usepackage[caption=false]{subfig}

\usepackage{xcolor}

\usepackage{microtype}

\aclfinalcopy 

\title{Where's the Question? A Multi-channel Deep Convolutional Neural Network for Question Identification in Textual Data}

\author{George Michalopoulos, Helen Chen,  Alexander Wong  \\
  University of Waterloo, \\  Waterloo, Canada \\
\texttt{\{gmichalo, helen.chen, alexander.wong\}@uwaterloo.ca} 
}
   
\date{}

\begin{document}
\maketitle
\begin{abstract}
In most clinical practice settings, there is no rigorous reviewing of the clinical documentation, resulting in inaccurate information captured in the patient medical records. The gold standard in clinical data capturing is achieved via ``expert-review", where clinicians can have a dialogue with a domain expert (reviewers) and ask them questions about data entry rules. Automatically identifying ``real questions" in these dialogues could uncover ambiguities or common problems in data capturing in a given clinical setting.

In this study, we proposed a novel multi-channel deep convolutional neural network architecture, namely Quest-CNN, for the purpose of separating real questions that expect  an answer (information or help) about an issue from sentences that are not questions, as well as from questions referring to an issue mentioned in a nearby sentence (e.g., \textit{can you clarify this?}), which we will refer as ``\textit{c-questions}". We conducted a comprehensive performance comparison analysis of the proposed multi-channel deep convolutional neural network against other deep  neural networks. Furthermore, we evaluated the performance of traditional rule-based and learning-based methods for detecting question sentences. The proposed Quest-CNN achieved the best F1 score both on a dataset of data entry-review dialogue in a dialysis care setting, and on a general domain dataset.
\end{abstract}

\section{Introduction}
In healthcare, real-world data (RWD) refers to patient data routinely collected during clinic visits, hospitalization, as well as patient-reported results. In recent years, RWD's volume has become enormous, and invaluable insights and real-world evidence can be generated from these datasets using the latest data processing and analytical techniques. However, RWD's quality remains one of the main challenges that prevent novel machine learning methods from being readily adopted in healthcare.  Therefore, creating data quality tools is of great importance in health care and health data sciences.  Erroneous data in healthcare systems could jeopardize a patient's clinical outcomes and affect the care provider's ability to optimize its performance.

Common data quality issues include missing critical information about medical history, wrong coding of a condition, and inconsistency in documentation across different care sites. Manual review by domain experts is the gold standard for achieving the highest data quality but is unattainable in regular care practices. Recent developments in the field of Natural Language Processing (NLP) has attracted great interest in the healthcare community since algorithms for identifying variables of interest and classification algorithm for diseases  have been recently developed \cite{Ji17}.

In this paper, we presented a novel model for the extraction of queries (questions) in a corpus of dialogue between data entry clinicians and expert reviewers in a multi-site dialysis environment.  
The main contributions of this work are:
\begin{enumerate*}[label=(\roman*)] 
\item To the best of our knowledge, we are the first to benchmark the performance of different rule and learning-based methods for the extraction of question sentences from logs of real-world (medical) systems by providing specific misclassification cases that emphasize the limitation of each technique.
\item  We proposed a new deep neural network architecture, namely Quest-CNN which unifies syntactic, semantic and statistical features and is capable of identifying real questions that expect an answer, questions referring to an issue mentioned in a nearby sentence (\textit{c-questions}) and   sentences that are not questions.
\item We examined the importance of the above mentioned features and we experimented extensively with different  state of the art deep learning models  in order to determine the best architecture for this particular task.
\item We investigated the effect of incorporating  domain knowledge on the performance of a model by examining whether   word embeddings and semantic features (that will be described in section \ref{features}) which are pre-trained in a domain-specific dataset rather than in a general dataset are more beneficial for the model.
\end{enumerate*}
Finally, in addition to evaluating our model's performance in a medical context, we also experimented in section \ref{expirement} with a general-domain dataset (questions in the Twitter social platform) to show our model's generalizability.

The rest of the paper is organized as follows. Related work is presented in section \ref{related}. The different question detection methods that will be examined,   are described in section \ref{question}. Section \ref{proposed} details the characteristics of the proposed multi-channel CNN model. Finally, the results of the experiments are reported in section \ref{expirement} and a conclusion and a plan for future work are given in section \ref{conclusion}.

\section{Related Work}
\label{related}
Different question-detection methods have  mainly been focused on the extraction of questions in social online settings (e.g. emails, Twitter) \cite{twi, Qinproceedings}. These methods can be classified into two categories: \begin{enumerate*}[label=(\roman*)] 
\item Rule based methods that make use of rules like 5W1H words (What, Who, Where, When, Why, How) or question marks (QM) to identify questions  
\item Learning-based methods, which train a classifier based on the patterns of sentences.
\end{enumerate*}

Recently, different deep-neural networks have achieved a state-of-the-art results in text classification.  In the \textbf{KIM-CNN} model\cite{kim-2014-convolutional} $t$ filters are applied to the concatenated word embeddings of each document in order to produce $t$ feature maps, which are  fed to a max pooling layer, in order to create a $t$-dimensional representation of the document.
In addition, in \cite{cnn-xml} the \textbf{XML-CNN} network  was introduced, where a dynamic max-pooling scheme and a hidden bottleneck layer were used to achieve a better representation of documents.  Another state-of-the art deep model is \textbf{Seq-CNN} \cite{johnson-zhang-2015-effective} where   each word is represented as a $\|V\|$-dimensional one-hot vector where $V$ is the vocabulary of the dataset and the concatenation of the word vectors are passed through a convolutional layer, followed by a special dynamic pooling layer. Furthermore, in \textbf{FastText} \cite{Joulin2016BagOT} the embedding of the words that  appear in a document were averaged to create a document representation. Finally, a comprehensive analysis of clinical-domain embedding methods is presented in \cite{khattak2019survey}.

\begin{figure*}
\centering
        \includegraphics[width=0.7\linewidth]{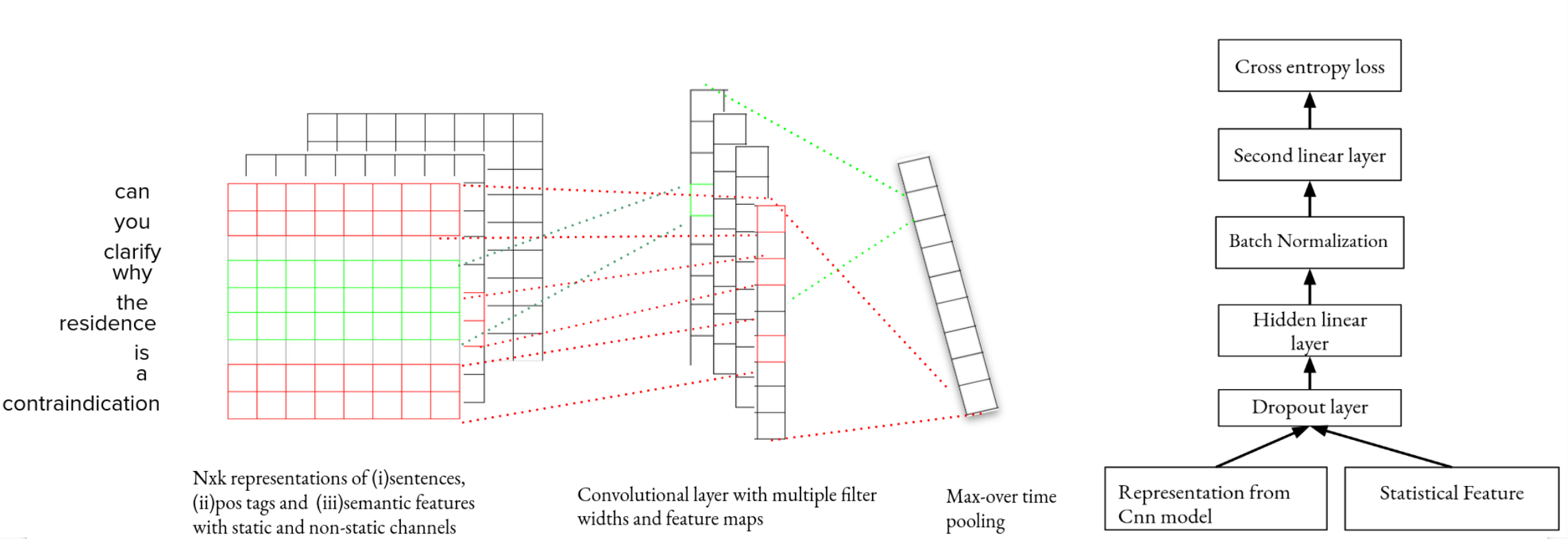}
    \caption{Proposed architecture of Quest-CNN with  the (three) feature channels for an example sentence}
    \label{fig:1}
\end{figure*}

\section{Question detection}
\label{question}
As mentioned above, the task of identifying sentences that contain questions and \textit{c-questions} can be broken into two sub-problems:  \begin{enumerate*}[label=(\roman*)]  \item Detecting question sentences from unstructured data (logs)\item  Correctly identifying the \textit{c-questions} and the real questions that require an answer. \end{enumerate*}

In order to create the corpus of candidate questions, we explored both rule and learning based approaches. We also compared the performance of each method on the task of identifying  questions on (medical) unstructured text and analyzed the most common misclassification cases.

\textbf{Rule-based Approach:}
In particular, we  employed the following rules: \begin{enumerate*}[label=(\roman*)] 
\item The last character of a sentence is a question mark
\item The rules that were introduced in \cite{Efron10} (e.g  I* [try*,like, need] to [find, know])
\item The sentence contains a 5W1H word
\item The refined 5W1H rules from \cite{twi}: The sentence contains a 5W1H word at its beginning or it contains auxiliary words  (e.g. ``what is" instead of what  or ``how does" instead of how)
\end{enumerate*}

\textbf{Learning-based Approach: }
These approaches learn specific patterns  that can be used to identify sentences that are questions. In this study, we evaluated the following methods:
\begin{enumerate*}[label=(\roman*)] 
\item  A Naive Bayes Classifier which was trained on the NPS Chat Corpus that consists of over 10,000 posts from instant messaging sessions \cite{Klein09}. As these posts have been labeled with  dialogue act types, such as \textit{``Statement"},\textit{``ynQuestion"}, we used the classifier  without any further training.
\item The syntactic parser from the Stanford CoreNLP Natural Language Processing Toolkit  \cite{manning-etal-2014-stanford} which provides a full  syntactic  analysis of an input sentence. The question-sentences were identified, by examining the syntactic structure of each sentence in the dataset.
\end{enumerate*}

\section{Quest-CNN Architecture}
\label{proposed}
In this section, we describe our proposed  model which has as input all the sentences that  the above question extraction method identified as questions and it distinguished between the sentences that are actual questions and \textit{c-questions}.

The architecture of our model is depicted in Figure \ref{fig:1}.
The key characteristic of  our model is its capability of unifying the syntactic properties of questions, the semantic  features of words by incorporating domain knowledge and statistical features (section \ref{features}) in a multi-channel CNN scenario.

Let $x_i \in \mathbb{R}^k$ be the $k$-dimensional word vector corresponding to the $i$-word of each sentence. Each sentence can be presented as the concatenation of the word embeddings of  its words. By using multiple $t$ convolutional filters   the model can capture multiple features for each sentence, as each  filter $u \in \mathbb{R}^{hk} $ will create one feature $c_i$ for each word region  $x_{i:i+h-1}$ (of size $h$):

\[c_i=f(u^T     \cdot x_{i:i+h-1} + b)\]
where $b \in \mathbb{R}$ is a bias term and $f$ is a nonlinear activation function  such as the rectified linear unit (ReLU).
By padding the end of the sentence, the model can produce a feature map $c_u=[c_1, c_2, ..., c_{n-h+1}]$ that is associated to a filter $u$,
 
where $n$ is the number of words of the largest sentence. In addition, by using $t$ filters (with different window sizes) the model  can create $t$ feature maps that are passed through a max pooling  operation in order to obtain one value $c_{max}$ for each filter. The output of the pooling layer is then concatenated with the vector of the statistical features of each sentence thus creating a richer representation of each sentence. Finally, the representation of the sentences are passed through two fully connected  layers and a softmax layer that outputs the probability distribution over the labels of the sentences.

Finally, our model consists of multiple channels that represent different features. 
Each filter $i$ is applied to all the channels and the feature map $c_i$ is calculated by adding the results in each channel. In the ablation study (section \ref{ablation}), we presented the performance of different variations of the model which consist  of one to three channels and  use (or not) the information of the statistical  features.

\subsection{Feature description}
\label{features}
Since the classification of the sentences is challenging due to their short length, we tried to utilize not only lexical features but also syntactical features,  statistical features and  semantic features using domain knowledge. In particular, we investigated the influence of the above mentioned features by creating multiple feature-channels for each sentence.

\textbf{Word representation: } 
The first channel consists of the $k$-dimensional word embeddings of the words of a sentence where   $x_i \in R^k$ corresponds to the $i$-th word in the current sentence. For the initialization of the word embeddings, we examine multiple options. We can either randomly initialize the vectors and then allow their modification during training or use pre-trained word embeddings that were trained in a much bigger corpus.

In the second case, we also investigated the effect of two popular pre-trained word embedding datasets. The first one is a general domain dataset, the Google News dataset that contains \textbf{300}-dimension word2vec embeddings that were trained on 100 billions words \cite{mikolov2013distributed}. The second one is the Multiparameter  Intelligent  Monitoring  in  Intensive  Care  (MIMIC) III dataset which is  a  publicly  available  ICU dataset \cite{mimiciii} that consists of medical data (e.g. laboratory tests, medical notes) collected between 2001 and 2012. Since a training course is mandatory in order to access it, there are no available word embeddings. Thus we trained our own embedding model that produced \textbf{300}-dimension word2vec embeddings  based on the \textbf{NOTEEVENTS} table which consists of 2,083,180 rows of patients notes. We believed that the comparison of the effectiveness of our model when it is provided with a large general domain dataset or a smaller specific domain dataset can be a useful guide not only to our work-case but for other NLP problems as well. Words that do not appear in the pre-trained vocabulary are randomly initialize using a uniform distribution in the range $[-\sqrt{\frac{3}{dim }}, +\sqrt{\frac{3}{dim }}]$  (where \textit{dim} is the dimension of word vector)\cite{Gustavo17}.

\textbf{POS-tagging word representation:}
Questions usually have specific syntactic structures and by including this channel we hoped to capture these meaningful syntactic features. By obtaining the POS tags of the words in each sentence,  we produced   embeddings of each  POS tag which has the same dimension as the word embeddings. The vectors of POS tags in the model can  either be initialized randomly  using a uniform distribution in the range $[-\sqrt{\frac{3}{dim }}, +\sqrt{\frac{3}{dim }}]$ 
and allow their modification during training or they can be created as one-hot vectors (vectors with all 0s and one 1 which indicates the existence of a specific POS tag word). By comparing these two representations we aimed to understand whether a richer representation of POS-tagging word can significantly improve the performance of the model.

\textbf{Semantic Features:}
Semantic features were also introduced in order to boost the model's performance by connecting different words that share a semantic meaning. In order to extract relevant clinical entities, we used the UMLS Metathesaurus, a large biomedical thesaurus, which organizes words by (medical) groups and links similar words \cite{UMLS}. The identification of the medical words and their UMLS groups was accomplished using the open-source Apache clinical Text Analysis and Knowledge Extraction System (cTakes)\cite{ctakesarticle}. We examined two different strategies for incorporating semantic features: \begin {enumerate*}[label=(\roman*)] 
\item Replacing the words  that appear in  the database  with their respective medical group name or
\item Creating a new channel  where, for each group, a new vector is created which was initialized either randomly or by using the pre-trained word2vec vectors as described above.
\end{enumerate*}

As our final goal for the model is to work in any domain, in the experiment section, we also presented our model's accuracy by extracting concepts from WordNet,  which is an extensive lexical database where words are grouped into sets of synonyms (synsets).    

Finally, four statistical features were included in the model namely the length of the sentence, the number of words, the number of capitalized words and the coverage of the vocabulary. These features were  introduced in \cite{Qinproceedings} for the identification of questions in the  Twitter platform.

\subsection{Regularization}
For regularization, we used ``embedding dropout", an idea that was introduced in language modeling in \cite{Merity2017RegularizingAO}, and performed dropout on entire word embeddings, thus removing some words in each training phrase. Although the ``embedding dropout" technique was used for the purpose of regularization  RNN-based models, we observed that it can work equally well in a CNN-based model as its main purpose is to make a model rely less on a small set of input words.

In addition, we applied dropout on the last two fully connected linear layers. The Dropout mechanism randomly sets a portion of hidden units to zero during training, thus preventing the co-adaptation of neurons \cite{Sri14}.

 \section{Experiments}
 \label{expirement}
In this section, we presented the result of an empirical evaluation on the question-detection methods that were described in section \ref{question} and our proposed model in section \ref{proposed}. In particular, we analyzed the performance of the rule  and learning based methods and we provided specific examples that highlighted their limitations. We also provided a comparison between different deep learning text classification methods that were described in section \ref{related} in order to show the efficiency of our proposed model. Finally, we conducted an ablation study that demonstrated the importance of the features (channels) for achieving  high-accuracy results.

\textbf{Datasets.} The main dataset used in this study is a review log consists of the query-answer dialogue in the Dialysis Measurement, Analysis and Reporting System (DMAR\textsuperscript{\textregistered}). DMAR\textsuperscript{\textregistered} is a web-based application that collects patient-level, clinical data within a renal program \cite{Bl13, Mat10} and its review module facilitates the ``query-answer” style of communication between the reviewer (Neurologist) and users (renal coordinators) during routine care process. A user would post a question to the reviewer only when she/he was unsure  about the correctness of patient data (for example \textit{why the hernia was considered absolute contraindications?’}). Therefore, the questions in this dialogue dataset provided a good indicator of data quality issues. The dataset used in this study was extracted from DMAR\textsuperscript{\textregistered} between 2013-2019. \footnote{REB\# 18-1604 approved by the Conjoint Health Research Ethics Board (CHREB), University of Calgary.}  This dataset offered a rare (in terms of quantity)  opportunity to examine a vast array of different questions types during medical data review processes over a fairly long period of time in a multi-sites real-world clinical setting. 
The annotations of the sentences were created by manually checking  each sentence, as a ground truth was non existent. Unfortunately, due to the sensitive nature of the dataset (contains medical information of patients), we cannot provide a link to a downloadable version of the data  without the approval of the research ethics boards.

\begin{table}[h]
\begin{center}
\begin{tabular}{|l|l|l|}
\hline
  & DMAR &Twitter\\\hline
Comments &  22125&  2462\\\hline
Questions-Sentences  & 4486&  2462\\\hline
Actual questions & 2575&  1262\\\hline
C-questions & 136&  -\\\hline
Comments with ?&  2722&  2462\\\hline
Comments  with 5W1H& 1780 &  836 \\\hline
A.N. of words & 15.0020&  11.4130\\\hline
A.N. of coverage of words & 0.0006& 0.0011\\\hline
A.N. of capitalize words&  2.3148&  2.1397\\\hline
A.N.  lengths& 85.4490&  66.3850\\
\hline
\end{tabular}
\end{center}
\caption{Statistics of the datasets; we use the acronym A.N for average number}
\label{tab:statistics_tab}
\end{table}

In addition, in order to test  our model in a general setting, we experiment with the Twitter dataset which was created and analyzed in \cite{Qinproceedings}. This dataset contains 2462 tweets which were annotated by two human annotators as questions (conveying an information  need) or non-questions. This dataset was used to evaluate the models  in the binary classification task to separate actual questions from sentences that do not require an answer. Table \ref{tab:statistics_tab} lists the statistics of both datasets.

For the evaluation of the question detection methods, the accuracy, recall and  F1 score (the harmonic mean between recall and precision) were reported.  For the evaluation of the deep-learning models, we reported the micro-averaged F1 score and the F1 score for the multi-class and binary-class dataset respectively on the testing and the validation set.

\subsection{Question Extraction}

The performance of different question extraction methods on the DMAR dataset was presented in Table \ref{tab:qep}.  The results showed that simple rules method, i.e. identifying question marks, had a medium performance.  The reason for the misclassification cases may  be due to the language patterns in casual conversational log, where people sometimes forgot to add a question mark (?) in a question (\textit{what symptoms did the patient present.}) or they used a question mark  to show irony (\textit{the procedure  has already begun so maybe an update next time?}),  or tried to be polite when they encouraged a person to take action (\textit{please see comments?}).

By adding the 5H1W rule, the recall could be boosted but the precision dropped significantly, due to the fact that many sentences could contain 5W1H but are not questions (e.g. \textit{just wait and see what happens}). Furthermore, we observed that by applying the refined rules introduced in \cite{twi}, the model could maintain almost the same recall, while achieving a significant improvement in precision. Finally, the results revealed that the rules in \cite{Efron10} had the overall best (F1) performance.  This indicated that even if these rules were made for a different context (twitter), they could be applied to other domains (e.g. medical).

\begin{table}[h]
\begin{center}
\begin{tabular}{|l|c|c|c|}
\hline
Methods &Prec.& Recall & F1 \\\hline\hline
Naive Bayes &0.772 &0.558 &0.648  \\\hline
Parsing algorithm &\textbf{0.930}&0.387&0.547\\\hline
QM & 0.862 &0.911&0.886\\\hline
QM \& 5W1H& 0.593 & \textbf{0.949} & 0.730 \\\hline
Rule 1 Li et al.\shortcite{twi}&  0.839& 0.916 & 0.876 \\\hline
Rule 2 Li et al.\shortcite{twi}& 0.826&0.916&0.869\\\hline
Rule 1,2 Li et al.\shortcite{twi}&0.861 &0.912&0.886\\\hline
Rule Efron et al.\shortcite{Efron10}& 0.862&0.918& \textbf{0.889}\\\hline
\hline
\end{tabular}
\end{center}
\caption{Precision, Recall and F1 scores for question identification methods; best results are \textbf{bolded}}
\label{tab:qep}
\end{table}

By investigating the performance of the learning-based approaches, we observed that the parsing algorithm could significantly improve the precision of question-detection but with low recall performance as it missed a large number of questions.  This is largely caused by the   irregular syntax patters used in real-life casual conversation (e.g. \textit{he is still sitting in my baseline?}). Finally, the Naive Bayes Classifier identified half of the questions with low precision which indicates that using transfer learning  (it is trained on the NPS Chat Corpus) cannot achieve a good performance especially in cases where the conversation is domain-specific, such as in a clinical setting or a electronic medical record system.
Thus, methods of this type  would need to be trained with domain-specific examples, which  are usually resource-intense.

It should be noted that we did not evaluate the  different question extraction methods on the Twitter dataset as it was constructed by tweets that were all ``potential'' questions (they all contain a question mark) and thus it cannot provide a fair comparison of the methods. To evaluate the QUEST-CNN model's ability to identify the actual question and the  \textit{c-questions}, we used the Twitter dataset and the dataset that contains all the sentences extracted by  all the question extraction methods from the DMAR dataset.

\subsection{Multichannel-CNN experiments}

In this section, we reported the evaluation of our proposed model in identifying questions and \textit{c-questions}. Firstly,  we presented a comparison of our model with the other deep-learning models that were described in section \ref{related}. In addition,  we also tested the performance of a bi-directional LSTM (BI-LSTM) model  that used as input all the features that we described in section \ref{features} where the last hidden state is fed to a fully-connected  softmax layer. Finally, we  compared our model  to logistic regression (LR) using an one-vs-rest multi-class objective and a support vector machine (SVM) model using a linear kernel. For the last two methods, we used the tf-idf measure of the words as features. 

In addition, we conducted an ablation analysis of our model. In summary, the baseline was a  \textbf{CNN-rand} model where  the word embeddings were randomly initialized but could be modified during training. We also provided the performance of our model when the word2vec embeddings were used (Google News or MIMIC) and we examined whether allowing further training of the embeddings (\textbf{CNN-non static}) can achieve better results than keeping them static(\textbf{CNN-static}). Finally, we examined whether the statistical, the POS-tag and the semantic features were meaningful features and which form of the features (as they were described 
in section \ref{features}) could achieve a better performance (\textbf{multi-channel} scenario).

All the experiments were performed with PyTorch 0.4.1. Scikit-learn 0.22.2  was used to implement the SVM, the LR model and the computation of the tf-idf vectors. All experiments were executed on  a Intel(R) Core(TM) i7-8665U CPU @ 1.90GHz with 15 GB RAM running Ubuntu 18.04.3 LTS.

\subsubsection{Hyperparameter Tuning}

\begin{table*}
  \begin{tabular}{l c c c c c c c c  }
  \hline
   &  \multicolumn{4}{c}{\underline{\hphantom{emptyemp} \textbf{DMAR} \hphantom{emptyem pemp}}} &  
  \multicolumn{4}{c}{\underline{\hphantom{emptyemp}\textbf{TWITTER}\hphantom{emptyempty}}}   \\
  Model & Test F1 &   Val. F1 & R.T  &   parameters &Test F1  &   Val. F1 &  R.T  &  parameters   \\  
    &   &    &  (sec.) &   &     &   & (sec.)  &   \\ \hline
    LR&  64.3$\pm$0.2& 65.0$\pm$0.4 & 0.5& 4091 &  53.2$\pm$0.3 & 57.4$\pm$0.5 &  0.3 & 7133  \\
SVM & 76.8$\pm$0.4&77.2$\pm$0.2 & 0.7 & 4091 & 61.2$\pm$0.3& 61.6 $\pm$0.4&  0.4 & 7133  \\ 
\hdashline
    KimCNN & 83.9$\pm$0.3 & 84.4$\pm$0.2 & 536& 2 578 371  & 61.7$\pm$0.4 & 62.7$\pm$0.5& 180  & 4 171 778 \\
    BILSTM & 82.9$\pm$0.2 & 83.3$\pm$0.2 & 168& 1 593 715 & 58.9$\pm$0.4 &  59.3$\pm$0.4 & 100  & 5 706 010  \\
    XMLCNN & 85.5$\pm$0.2& 86.0$\pm$0.3& 319& 2 209 953 & 61.1$\pm$0.5 &61.6$\pm$0.5& 158 & 3 80 3775   \\
    SeqCNN & 84.2$\pm$0.1& 84.4$\pm$0.1&  3191 & 1 243 735 & 64.7 $\pm$ 0.2 & 65.1$\pm$ 0.2&   3142& 1 242 701   \\
    FastText & 77.8$\pm$0.1& 79.9$\pm$0.1&\textbf{45}& 1 200 003& 63.0 $\pm$0.3  & 63.6 $\pm$0.4& \textbf{36}  & 2 793 602 \\
    QuestCNN & \textbf{86.9$\pm$0.2} & \textbf{87.4$\pm$ 0.3} & 594 & 2 988 183 & \textbf{ 65.5$\pm$0.2} & \textbf{ 66.7 $\pm$0.3}& 399 & 7 100 486 

  \end{tabular}
  \caption{Results of mean $\pm$ standard deviation of five runs from each model on the test and the validation test;  average running time (R.T.) and number of trainable parameters are also provided for each model. The number of parameters is different between datasets as we included the embeddings vectors that are fine-tuned for each dataset;   all models used the same (MIMIC) embedding; best values are \textbf{bolded} }
  \label{tab:2}
\end{table*}

In order to address the concerns of reproducibility of the NLP community \cite{showyourwork},  we provided the search strategy and the bound for each hyperparameter as  follows: the batch size was set between 32 and 64; the dropout embedding and the learning rate were from  a uniform distribution on the interval $[0,0.5]$,  $[0,0.2]$ and $[1\mathrm{e}$-$6,1\mathrm{e}$-$1]$  respectively; the filter sizes of the models were between (2,3,4), (3,4,5)  and (2,4,8); the feature map, the dynamic pool, hidden linear layer and the hidden size of the BI-LSTM were from a discrete uniform distribution on the interval $[100,200]$, $[2,6]$, $[40, 100]$, $[25,75]$ respectively; the SEQ-CNN filter size was between the values 2,3,4 and feature map from a discrete uniform distribution on the interval $[800, 1200]$. The best assignment was chosen based on the F1 value  achieved on the DMAR validation set after $50$  search trials.  We did not perform any other dataset-specific tuning. An implementation of all the models is downloadable in our github repository \footnote{\url{https://github.com/gmichalo/question_identification_on_medical_logs}}.

In the interest of providing a fair comparison we tuned the hyperparameters of each model. To train XML-CNN, we selected a dynamic pooling window of length 3, a learning rate $5\mathrm{e}$-$5$, a batch size of 32, feature map 128, hidden linear layer of size 77, filter sizes of (3,4,5) and dropout of 0.338. For Kim-CNN, we used batch size 64, filter sizes of (2,4,8), feature map of 164, learning rate 0.003 and dropout 0.077. For FastText, we used a batch size of 64 with learning rate 0.051. For training the Bi-LSTM, we chose a batch size of 64, learning rate 0.004, dropout of 0.056 and hidden size of 50. For SEQ-CNN, we used a batch size of 32, learning rate 0.084, filter size of 4, feature map 1033 and dropout 0.215. Finally, for the SVM and the LR model, we chose the default hyperpameters of the scikit-learn library.

For the QuestCNN, we used a batch size of 32, a learning rate 0.012, a feature map of 160, filter sizes of (3,4,5), a hidden linear layer of size 96, a dropout rate of 0.164 and an embedding dropout of 0.016. In addition, we applied spatial batch normalization to the output of each channel of our model and batch normalization to the output of the hidden linear layer which is a technique for normalizing layer inputs. This technique has been shown  to accelerate the training of different deep learning models \cite{Iof15}. The Adam optimizer \cite{Adam}  was chosen, with cross-entropy loss,  as our optimization objective. Finally, each model was trained for $30$ epochs.

\begin{figure}[h]
\centering
 
        \includegraphics[scale=0.08]{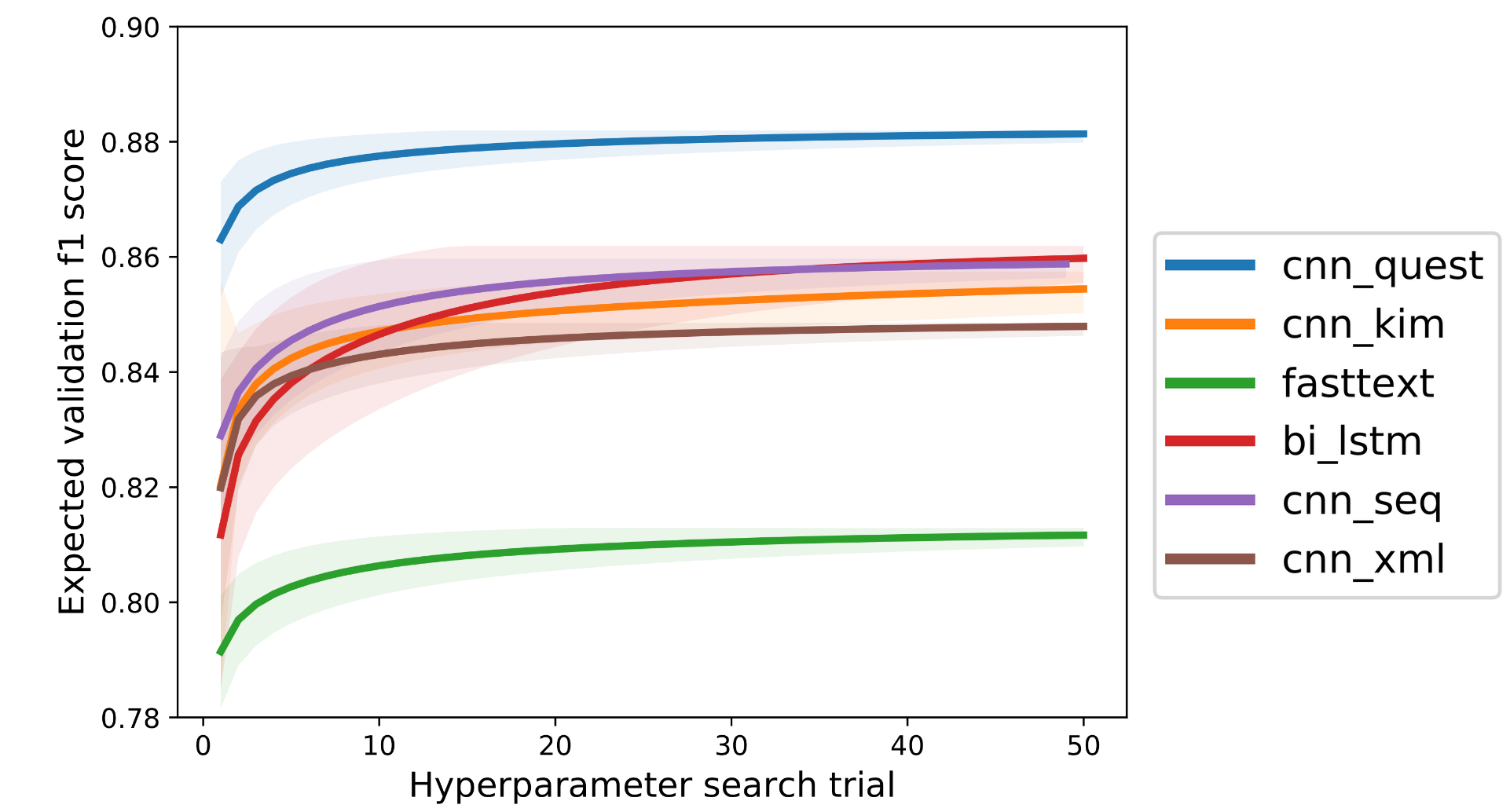}
    \caption{Expected validation performance of all the deep learning models over 50 hyperparameter trials. The standard deviation of the expected performance is represented as the shaded area in the graph}
    \label{fig:expected1}
\end{figure}

Furthermore, in Figure \ref{fig:expected1}, we provided the validation performance of the models based on the technique in \cite{showyourwork} that measures the mean and variance of the performance of a model  as a function of the number of hyperparameter trials.
Finally, as the datasets do not have a standard dev set, we split the dataset to $80\%$ training, $10\%$ validation and $10\%$ testing. In order to present more robust results,  we ran our model on five different seeds and we provided the average  scores for the testing and the validation set.

\subsubsection{Deep Learning Model comparison}

The  mean and standard deviation (SD) of the scores for our model and the other competing models on the task of identifying questions are reported in Table \ref{tab:2}. QUEST-CNN was shown to have achieved the best F1 score for both datasets ($86.9\%$ and $65.5\%$). This is due to the fact that the model utilizes information from all the features (syntactic, semantic, statistical) and uses  regularization and normalization techniques. In section \ref{ablation}, we analyzed in detail the effect of each feature. The FastText model performed the poorest on DMAR ($77.8\%$) as this model didn't consider the word order of the sentence  but it required the least amount of running time. In addition, Kim-CNN, XML-CNN and Seq-CNN architectures had similar performances on the DMAR dataset ($83.9\%$, $85.5\%$ $84.2\%$) but on the twitter dataset Seq-CNN had the second best performance (after Quest-CNN) with $64.2\%$. However, the running time of  the Seq-CNN is the largest by a considerable margin. Furthermore, the Bi-LSTM model which utilized all the features, achieved a non-optimal performance, which confirmed our hypothesis that a CNN-architecture is more suitable for the task of question identification.   Also, it should be noted that SVM and LR achieved a decent performance. SVM in particular, achieved a similar performance to the FastText model on the DMAR dataset and even surpassed the BI-LSTM on the Twitter dataset.

\begin{figure*}
\centering

        \includegraphics[ width=0.94\linewidth]{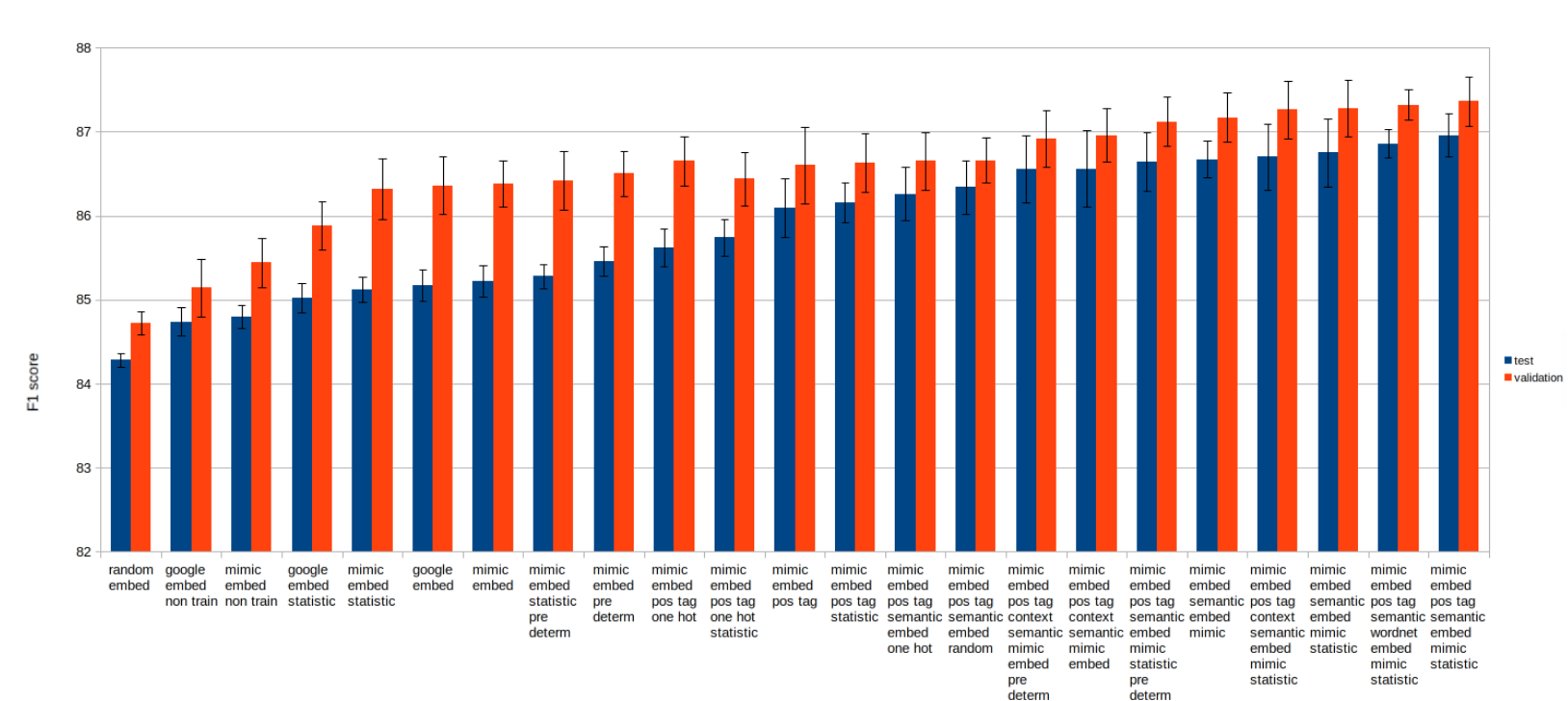}
    \caption{F1 scores of the Ablation study for the Quest-CNN in the DMAR dataset;  the word context means that we did not create a new channel but we replaced the words in the sentence with their semantic group name.}
    \label{fig:ablation}
\end{figure*}

\subsubsection{Ablation Study}
\label{ablation}

The comparison of the performance of different variants of our model is presented in Figure \ref{fig:ablation}. Firstly,  even our baseline model (CNN-rand) performed better than classic machine learning models (LR and SVM). In addition, by  using pre-trained vectors a static model could achieve a better performance than CNN-rand. Further fine-tuning of the pre-trained models could improve the performance even more. Finally, the multi-channel model achieved a better performance than the single channel model in every case, as the best performance was achieved by using all the 3 feature types and the statistical features. We observed that the strategy of creating a new channel for the semantic features achieved a better performance than replacing the words with their respective medical group name.  This provided evidence that a richer representation of the sentences could help the model to make more accurate evaluations.

\subsubsection{Using Domain Knowledge}

By comparing the behavior of our model when it utilizes  pre-trained word embeddings from a general (Google) or a specific (Mimic) domain,
(for the word embedding channel and for the semantic feature channel)
we observed that the MIMIC III pre-trained embeddings have a more positive effect on the behavior of the model.   In addition, we experimented with the creation of the semantic channel by either using  medical group names (from UMLS, a medical Metathesaurus) or concepts from WordNet (a general-domain  database). For WordNet, as the  assignment  of  words  to  concepts  can be  ambiguous, we considered only the first concept. This choice was based on the assumption that WordNet returns a list of concepts where the most common meaning is listed first \cite{Elbe08}. Similar to the embeddings case, we observed that concepts from a specific domain are more beneficial for the performance of the model  (Figure \ref{fig:ablation}).
These experiments indicated that when applying a classification model on documents of a specific domain (i.e medical), exploiting domain knowledge from a  dataset is more advantageous than using general knowledge from a larger dataset. Finally, our experiments showed that using one-hot representation was generally weaker than a word embedding representation, even if the embeddings were initialized randomly and were allowed to be further updated during the training phase of the model.

\subsubsection{Using Pre-determined Characteristics}

Finally, we examined whether adding information from the question extraction method could improve the performance of the model. Specifically, for each sentence, a $n$-vector was created where $n$ was the number of question extraction methods. For each position $i$ was set to 1 if the $i$ method classified the sentence as a question. Otherwise, it was set to 0. This vector was concatenated with the output of the max-pooling layer of the Quest-CNN model (like the statistical features) and then fed to the last fully connected layers in our model. By comparing the performance of the model when these characteristics were available to it (Figure \ref{fig:ablation}),  we could observe that our model was capable of identifying these characteristics on its own, and  any external guidance would not further improve the performance of the model.

\section{Conclusion and Future work}
\label{conclusion}
In this paper, we have provided an analysis of the performance of existing methods for question extraction with real-world misclassification examples that showed the weak point of each method. Furthermore, we have proposed a novel approach for the automatic identification of real questions   and \textit{c-questions}. We have also shown empirically that the proposed architecture of unifying syntactic, semantic and statistical features achieved a state-of-the-art F1 score for this particular task. Finally, we have presented the relevance of exploiting domain knowledge in the overall performance of a model.

We are in the process of obtaining access to datasets from different application contexts in order to examine the generalizability of our model. As for future work, we plan to extend our work by calculating the similarity of questions in order to create groups of questions that represent the most impactful ``problems'' of a given application environment. Finally, we plan to compare our model with recent  language   representation  models like the BERT model  in \cite{devlin2018bert} both for the task of question identification and for the task of creating the above mentioned ``problem'' groups.

\bibliography{emnlp2020}
\bibliographystyle{acl_natbib}

\clearpage

\appendix

\appendixpage

\section{UMLS  Metathesaurus}
\begin{figure}[h]
\centering
        \includegraphics[totalheight=3.5cm]{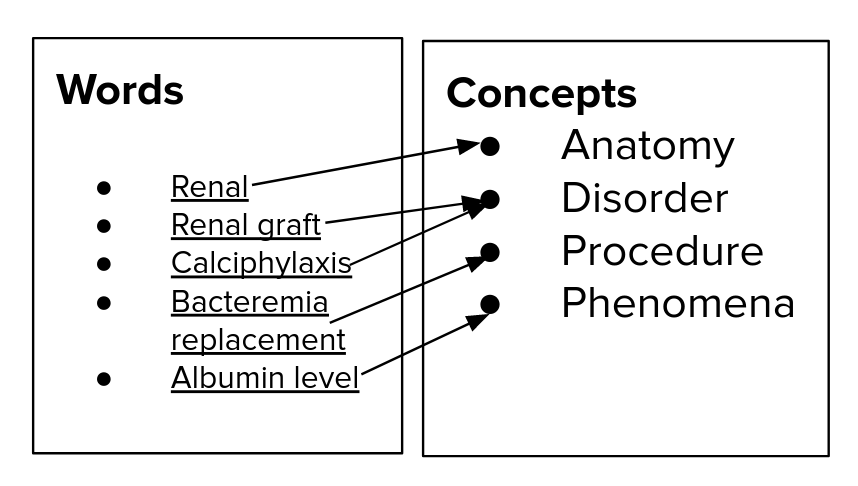}
    \caption{Example of mapping words to medical group}
    \label{fig:concept}
\end{figure}

As we described in the paper, for the creation of the semantic features of our model, relevant medical group names from the UMLS Metathesaurus were extracted. 
An example of mapping words to medical group names can be observed in Figure \ref{fig:concept} where different  words that have a medical significance can be categorized   as a medical procedure, a disorder or an anatomy term.

We also provide a complete list of  all the Semantic Group Names   and  Semantic Type Names in Table \ref{tab:sem} that our semantic feature model can choose from. In our experiment, we chose to only use the more general  Semantic Group Names, since sentences are generally short in length, so we wanted to have a dense connection between different words that are semantically similar as possible. Finally, it should be noted that UMLS also contained additional Semantic groups (like Occupations, Organizations), which we decided to remove from the process of  creation of semantic features, as these groups are irrelevant to the domain of the testing dataset. However,  these medical groups did not influence the general performance of our model.

\begin{table}[h!]
  
\begin{tabular}{|l|c| }
 \hline
 Group Name &   Semantic Type Name \\\hline\hline

Anatomy&Anatomical Structure\\
Anatomy&Body Location or Region\\
Anatomy&Body Part, Organ, or Organ Component\\
Anatomy&Body Space or Junction\\
Anatomy&Body Substance\\
Anatomy&Body System\\
Anatomy&Cell\\
Anatomy&Cell Component\\
Anatomy&Embryonic Structure\\
Anatomy&Fully Formed Anatomical Structure\\
Anatomy&Tissue\\
Disorders&Anatomical Abnormality\\
Disorders&Cell or Molecular Dysfunction\\
Disorders&Congenital Abnormality\\
Disorders&Disease or Syndrome\\
Disorders&Experimental Model of Disease\\
Disorders&Finding\\
Disorders&Injury or Poisoning\\
Disorders&Mental or Behavioral Dysfunction\\
Disorders&Neoplastic Process\\
Disorders&Pathologic Function\\
Disorders&Sign or Symptom\\
Phenomena&Biologic Function\\
Phenomena&Environmental Effect of Humans\\
Phenomena&Human-caused Phenomenon or Process\\
Phenomena&Laboratory or Test Result\\
Phenomena&Natural Phenomenon or Process\\
Phenomena&Phenomenon or Process\\
Procedures&Diagnostic Procedure\\
Procedures&Educational Activity\\
Procedures&Health Care Activity\\
Procedures&Laboratory Procedure\\
Procedures&Molecular Biology Research Technique\\
Procedures&Research Activity\\
Procedures&Therapeutic or Preventive Procedure\\ \hline
\end{tabular}
\caption{ The Semantic Group Names   and the  Semantic Type Names from UMLS library that were used for the semantic features.}
\label{tab:sem}
\end{table}

\section{C-question Format}

In this paper, we introduced the notion of \textit{c-question}, which is a question referring to an issue mentioned in a nearby  sentence. The annotations of the sentences were created by manually checking  each sentence and we adhered to the following rules:

A sentence can be classified as a \textit{c-question} if:
\begin{itemize}
    \item It can be classified as a real  question  i.e it expects an  answer  (information  or  help)  about  an  issue.
    \item The issue of the question should not be recognizable from the sentence of the question. Usually the question would used  a demonstrative pronoun (can you clarify this?).
    \item The issue of the question should only be identified in a nearby sentence.
\end{itemize}

\begin{table}[h]
\begin{tabular}{ |c|c|c| } 
 \hline
  & question & c-question \\\hline
A.N. of coverage&   $6.10^{-4}$ & $2.10^{-4}$ \\ \hline
A.N. of length &  83.1 & 24.8 \\ \hline
A.N. of words &  14.4 & 4.9 \\ \hline
A.N. of dem. pronoun &  0.3 & 0.3 \\ \hline
 
 \hline
\end{tabular}
 \caption{Statistics of  questions and c-questions; we use the acronym A.N for average number and dem. for demonstrative pronoun.}
\label{tab:statistics_tab_c}
\end{table}

Finally, in Table \ref{tab:statistics_tab_c}, we list the statistics of the sentences that were classified as question-sentences and the \textit{c-questions} in order to make the  distinguishment between these categories more understandable. For example,  we can observe that even if the average number of words of questions is 2.9 higher than the average number of words in \textit{c-questions}, the average number of demonstrative pronouns is the same for both categories.

\end{document}